\documentclass[10pt,twocolumn,letterpaper]{article}

\usepackage[pagenumbers]{cvpr} 

\definecolor{cvprblue}{rgb}{0.21,0.49,0.74}
\usepackage[pagebackref,breaklinks,colorlinks,allcolors=cvprblue]{hyperref}

\def\paperID{11034} 
\def\confName{CVPR}
\def\confYear{2026}

\title{MV2UV: Generating High-quality UV Texture Maps with Multiview Prompts}

\author{
    Zheng Zhang$^{1}$\thanks{Equal contribution} \qquad 
    Qinchuan Zhang$^{1\ast}$ \qquad 
    Yuteng Ye$^{1}$ \qquad 
    Zhi Chen$^{1}$ \\
    Penglei Ji$^{1}$ \qquad 
    Mengfei Li$^{2}$ \qquad 
    Wenxiao Zhang$^{1}$ \qquad 
    Yuan Liu$^{2}$\thanks{Corresponding author} \\
    \and
    $^{1}$Hisilicon Linx Lab, Huawei \qquad \quad $^{2}$Hong Kong University of Science and Technology \\
    {\tt \{zhangzheng119, zhangqinchuan3\}@huawei.com} \qquad {\tt yuanly@ust.hk}
    \vspace{-0.1em} 
}

\let\oldtwocolumn\twocolumn
\renewcommand\twocolumn[1][]{%
    \oldtwocolumn[{#1}{
    
\begin{center}
\vspace{-1.5em}
\includegraphics[width=0.95\textwidth]{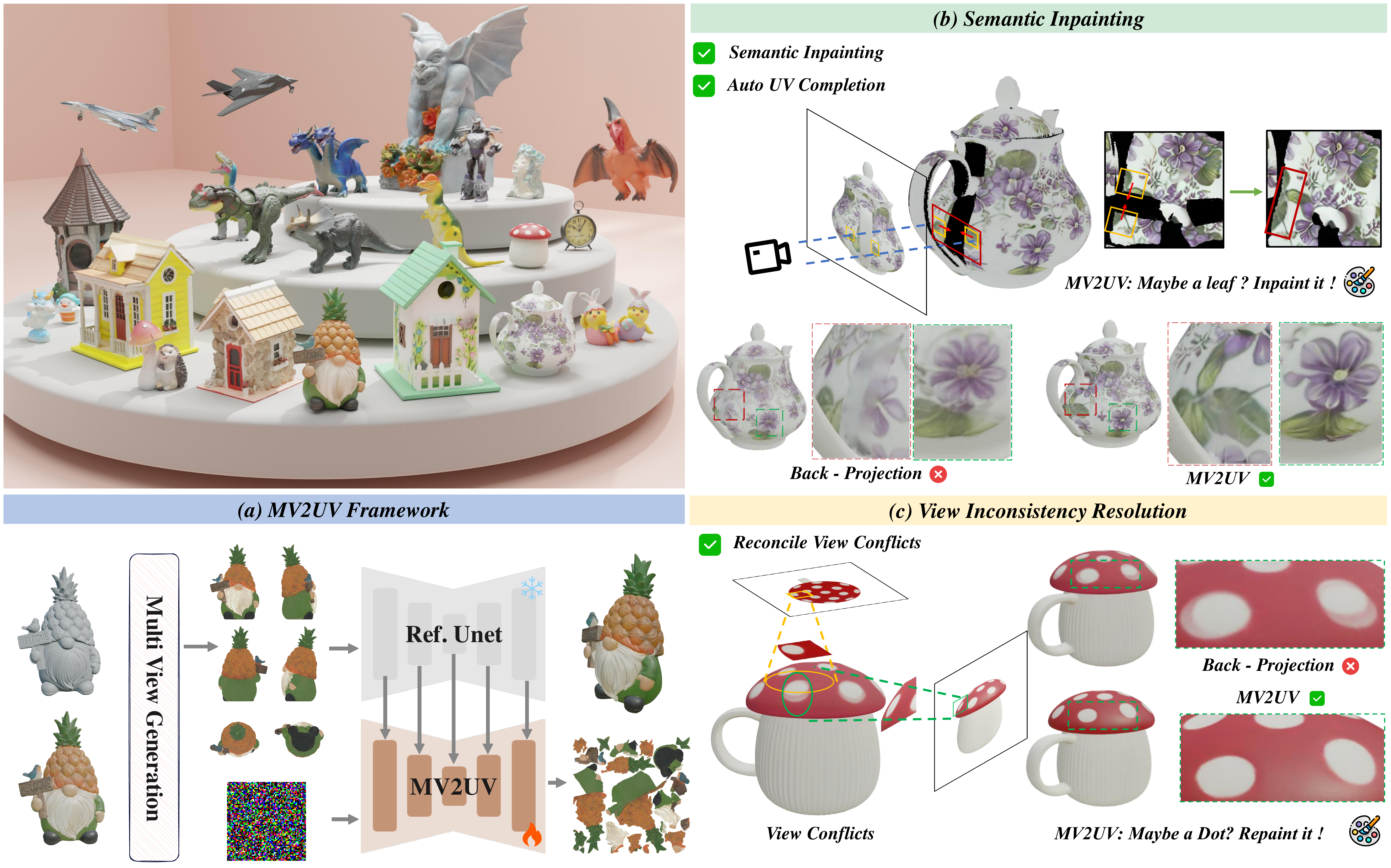}
    \captionof{figure}{Our \textbf{MV2UV} is a novel two-stage method that treats generated multi-view (MV) images as semantic prompts to guide texture generation on the UV map. \textbf{The left top} shows a set of textured 3D meshes generated by our method. \textbf{(a)} MV2UV enables the combination of the image diffusion prior with the UV inpainting. \textbf{(b)} MV2UV uses MV images as prompts for semantically consistent inpainting. \textbf{(c)} Though MV images show inconsistency, MV2UV enables automatically resolve the inconsistency in the UV generation process.
    }
    \label{fig:teaser}
\end{center}
    }]
}

\begin{document}

\maketitle
\begin{abstract}
Generating high-quality textures for 3D assets is a challenging task. 
Existing multiview texture generation methods suffer from the multiview inconsistency and missing textures on unseen parts, while UV inpainting texture methods do not generalize well due to insufficient UV data and cannot well utilize 2D image diffusion priors. In this paper, we propose a new method called MV2UV that combines 2D generative priors from multiview generation and the inpainting ability of UV refinement to get high-quality texture maps. Our key idea is to adopt a UV space generative model that simultaneously inpaints unseen parts of multiview images while resolving the inconsistency of multiview images. Experiments show that our method enables a better texture generation quality than existing methods, especially in unseen occluded and multiview-inconsistent parts.
\end{abstract}    
\section{Introduction}
\label{sec:intro}

Texture mapping for 3D assets is a critical link in 3D content production pipelines, as it determines the visual quality of 3D assets in fields like game development, virtual reality, and digital twins. Traditionally, high-quality texture creation relies heavily on manual work by professional artists, which limits the scalability of 3D content production.

Most existing texture generation methods fall into two categories, both of which struggle to balance realism and efficiency in practical 3D content production. The first category adopts a multiview-based method (e.g., Paint3D~\cite{zeng2024paint3d}, MaterialMVP~\cite{huang2025material}, MV-Adapter~\cite{huang2025mv}, RoomTex~\cite{wang2024roomtex}): textures are generated from multiple views and then projected onto the UV map. However, this approach suffers from two critical flaws: (1) multiview inconsistency, where misalignments between views lead to blurry or conflicting texture regions; (2) poor handling of unseen/occluded areas, which can only be filled via smooth extrapolation~\cite{cheng2025mvpaint} or supplemented by basic UV-space inpainting that lacks semantic grounding~\cite{huo2024texgen,zeng2024paint3d}. The second category generates textures directly on the UV map (e.g., TEXGen~\cite{huo2024texgen}), but it lacks sufficient priors in the UV domain (since UV coordinates do not inherently encode 3D spatial or semantic information), resulting in textures that often fail to match the object’s structural logic. These limitations collectively compromise texture realism and hinder the scalability of 3D asset production.

To address these challenges, we propose a novel two-stage framework that combines the strengths of multiview semantics and UV-based generation while mitigating their weaknesses. Specifically, as shown in Fig.~\ref{fig:teaser} (a), we first generate multiview (MV) images using multiview diffusion. Instead of directly back-projecting these multiview images to the UV map, we treat them as semantic prompts to guide texture generation on the UV map. This design offers two key advantages over prior work: (1) Compared to conventional multiview generation methods, our framework enables automatic completion of occluded UV regions and avoids direct projection-induced inconsistencies. Instead, the network learns to autonomously reconcile potential view conflicts during UV-based generation. (2) Compared to direct UV inpainting/generation methods, our approach injects rich multiview priors into the UV domain. Unlike UV-only methods that lack semantic context, our MV prompts provide explicit structural and textural cues (e.g., surface patterns from multiple angles). 

A critical challenge in our framework is designing an effective multiview prompting technique that can reliably link multiview semantics to the UV map. 
Our core insight is that when conducting cross-attention between the UV map and the prompting MV images, we will use their pixel-wise aligned 3D coordinates map (XYZ) as the positional encoding to guide the attention to attend to meaningful regions. This design leverages the fact that XYZ coordinates uniquely map each pixel (in both MVs and the UV map) to its physical location on the 3D object, enabling precise cross-modal correspondence. A pixel in the UV map will dynamically attend to all MV regions that share the same XYZ coordinates. This XYZ-based attention brings two key benefits: (1) Superior conflict resolution: Rather than direct projection/averaging projections, which often blur details or produce incorrect textures with multiview inconsistency, the network can adaptively resolve the inconsistency to generate correct and sharp details on the UV map, as shown in Fig.~\ref{fig:teaser} (c). (2) Enhanced semantic inpainting: For UV regions invisible to MV images, the attention mechanism allows the model to pull semantic information from adjacent visible MV regions, ensuring inpainted textures align with the object’s overall structure, as shown in Fig.~\ref{fig:teaser} (b).



We evaluate our framework on the DTC~\cite{dong2025digital} and GSO~\cite{downs2022google} datasets covering diverse 3D objects (e.g., rigid furniture, deformable fabrics) and lighting conditions. Quantitative and qualitative results demonstrate its superiority: (1) Compared to UV-based generation methods (e.g., TEXGen~\cite{huo2024texgen}), our textures exhibit more plausible structural and textural details, while achieving FID improvements of 50.8 on GSO and 14.7 on the DTC dataset. (2) Compared to multiview based generation methods (e.g., Hunyuan3D 2.1~\cite{hunyuan3d2025hunyuan3d}), our framework reduces inconsistency artifacts with significant FID improvements and 2 times better completion quality for occluded regions.
\section{Related Work}
Recent advances in texture generation for 3D assets can be broadly categorized into two major paradigms: \textit{optimization-based} and \textit{feed-forward} approaches. The former typically relies on iterative optimization guided by pre-trained 2D diffusion models, while the latter directly produces coherent, view-consistent textures through an end-to-end generative framework.

\subsection{Optimization-based Texture Generation}

Optimization-based texture generation approaches ~\cite{poole2022dreamfusion,metzer2023latent,wang2023prolificdreamer,lin2023magic3d, zhang2024texpainter,yeh2024texturedreamer,chen2023fantasia3d,jiang2025flexitex, earle2024dreamcraft, wu2024texro} employ pre-trained diffusion models as supervisory signals, progressively refining the texture through iterative optimization at inference time. Broadly speaking, these methods diverge along two paths: one refines the entire 3D neural representation as a unified whole—its textures naturally emerging from the learned structure, while the other concentrates solely on optimizing texture maps upon a predefined geometry.

DreamFusion~\cite{poole2022dreamfusion} introduced the Score Distillation Sampling (SDS) approach, leveraging 2D supervision from diffusion models to optimize a 3D NeRF representations without the need for any 3D training data. ProlificDreamer~\cite{wang2023prolificdreamer} introduces Variational Score Distillation (VSD), which improves SDS by modeling the distribution of 3D representations instead of a single deterministic sample, enabling more diverse texutre synthesis. DreamMat~\cite{zhang2024dreammat} tackles baked-in shadow artifacts in PBR material generation with a light-aware 2D diffusion model that jointly optimizes for multiple high-quality material maps.

The second category of methods ~\cite{zhang2023dreamface, liao2024tada, zhang2024teca} directly generates view-consistent texture maps for a specified mesh through optimization. These methods extend SDS to optimize human UV textures under a unified human parametric model SMPL-X. DreamPolish~\cite{cheng2024dreampolish} and Paint-it ~\cite{youwang2024paint} further generalize texture optimization to 3D models of arbitrary topology and diverse categories.

Despite the advantage of requiring no large-scale 3D datasets, optimization-based frameworks remain computationally expensive at inference, and are further hindered by common artifacts such as the Janus problem.

\subsection{Feed-forward Texture Generation}

Feed-forward approaches can be roughly divided into three categories. 1) View-by-view approaches ~\cite{zeng2024paint3d, huang2025material, richardson2023texture} employ an iterative generation process that synthesizes textures across multiple viewpoints in sequence, ensuring that subsequent textures remain consistent with previously generated ones. These methods hinder joint optimization across multiple views and are computationally inefficient, as the inherently sequential nature of their process prohibits parallel generation. 2) Multiview approaches ~\cite{feng2025romantex, zhao2025hunyuan3d, huang2025mv, xu2025wondertex, wang2024crm,li2025step1x, bensadoun2024meta} simultaneously generate multiple images of a 3D object from different viewpoints, then back-project them onto the UV map to obtain the final texture. Recently, several multiview generation-based methods ~\cite{he2025materialmvp, zhang2024clay, li2024idarb, he2025neural, zhu2025muma} have extended texture generation to the generation of PBR material components. By leveraging cross-view attention to share features across views, the generated textures exhibit stronger view consistency and can be produced more efficiently. Nevertheless, this approach still struggles with partial texture loss when generating occluded regions of self-occluding objects. To address the issue of incomplete coverage, several methods have employed inpainting techniques either in UV space ~\cite{zeng2024paint3d, huang2025material,yan2025flexpainter} or 3D space ~\cite{liang2025unitex, cheng2025mvpaint,georgiou2025im2surftex}, thereby further enhancing the completeness of the generated texture maps. 3) UV-based approaches tackle occlusion by synthesizing full texture map ~\cite{huo2024texgen, yuan2025seqtex} or tiling material ~\cite{xin2024dreampbr, chen2024dtdmat, luo2024correlation, zhang2024mapa} directly in UV space. However, directly performing generation work on UV texture maps fails to establish a connection between UV and view space, making it difficult for reference images to provide sufficient prior information to guide the generation of texture maps.

\begin{figure*}
    \centering
    \includegraphics[width=0.98\linewidth]{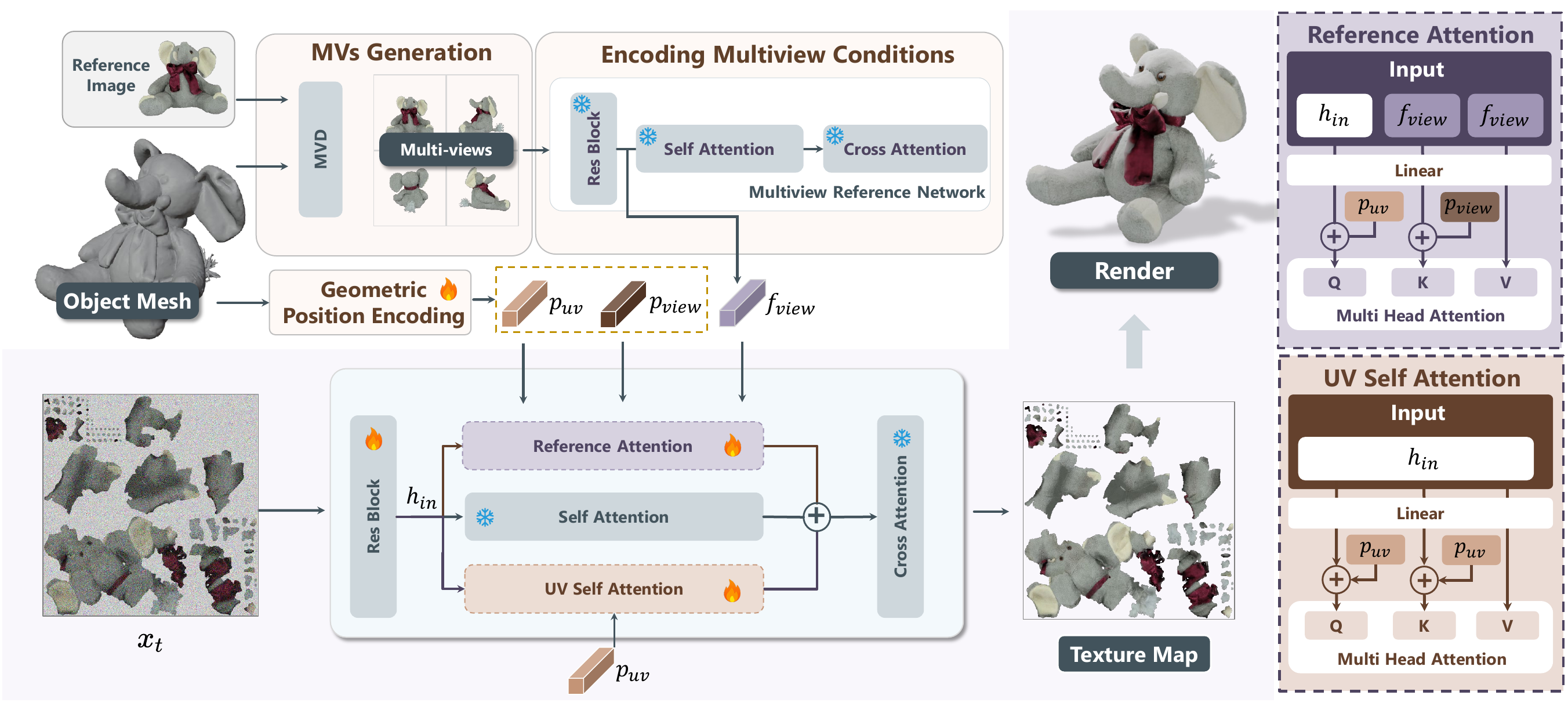}
    \caption{The Overview of our MV2UV framework. We first generate multiview images (MVs) using multiview diffusion (MVD). Then, we treat the generated MVs as semantic prompts to guide texture generation on the UV map. The generation process is based on a diffusion U-Net, which primarily consists of Self Attention, UV Self Attention, and Reference Attention modules.}
    \label{fig:pipeline}
\end{figure*}

\section{Method}
\label{sec:formatting}


Given an untextured mesh with known UV mapping and an image illustrating the texture of the object, our framework is designed to generate texture maps for this mesh, which can be combined with arbitrary existing image-to-3D generation pipelines. Our framework combines a multiview diffusion model with a UV refinement diffusion model to generate high-quality texture maps. In the following, we first introduce the high-level idea of our framework.

\subsection{Overview} 
The overview of our pipeline is given in Fig.~\ref{fig:pipeline}. From the given input images, we first generate multiview images using multiview diffusion. Then, these generated multiview images are utilized as image prompts in a UV texture map diffusion generative model to generate a complete UV texture map for the object mesh.

\textbf{Discussion}. As discussed in the introduction (Sec.~\ref{sec:intro}), a straightforward approach to leveraging multiview images for texture generation is to bake-project all views directly onto the mesh to construct the target texture map. However, inherent inconsistency among the generated multiview images often leads to conflicts during back-projection, resulting in blurry or visually inconsistent regions in the final texture. Furthermore, occluded areas invisible to the multiview cameras can only be filled with uniform, low-semantic colors via simplistic inpainting, which inevitably leads to the loss of fine-grained details in these regions. To address both limitations, our framework abandons direct back-projection of multiview images onto the mesh. Instead, we utilize these images as semantic prompts to guide texture generation on the UV map, enabling the model to inpaint high-fidelity textures for unseen/occluded regions while adaptively resolving conflicts of multiview inconsistency.

\textbf{Multiview Diffusion}. 
Multiview diffusion models~\cite{liu2023zero, shi2023mvdream, shi2023zero123++, hu2024mvd} extend image diffusion frameworks to jointly generate view-consistent images across multiple camera viewpoints, ensuring geometric and appearance coherence. For our framework, we employ MV-Adapter as the multiview generation module. Its parallel attention architecture—integrating multiview attention and image cross-attention—enhances pre-trained text-to-image models for consistent multiview generation under diverse conditions. In the following, we introduce how to utilize the generated multiview image prompts in the UV texture map generation.

\subsection{Image-prompted UV generation}
Given the multiview generations from the multiview diffusion model, we adopt a latent image diffusion model to directly generate the texture map on the UV map, using the multiview generations as prompts. Specifically, our backbone network is finetuned from the Stable Diffusion XL (SDXL)~\cite{podell2023sdxlimprovinglatentdiffusion} model with frozen self attention and cross attention modules. In the following, we introduce how we adapt the original SDXL model to a UV generation model, in terms of utilizing multiview images as conditions (Sec.~\ref{subsec:mv}), adding geometric positional encoding as guidance (Sec.~\ref{subsec:positionembedding}), and adding a new UV self attention (Sec.~\ref{subsec:uv}). These new designs are summarized in Sec.~\ref{subsec:summary}.


\begin{figure}[h]
    \centering
    \includegraphics[width=0.98\linewidth]{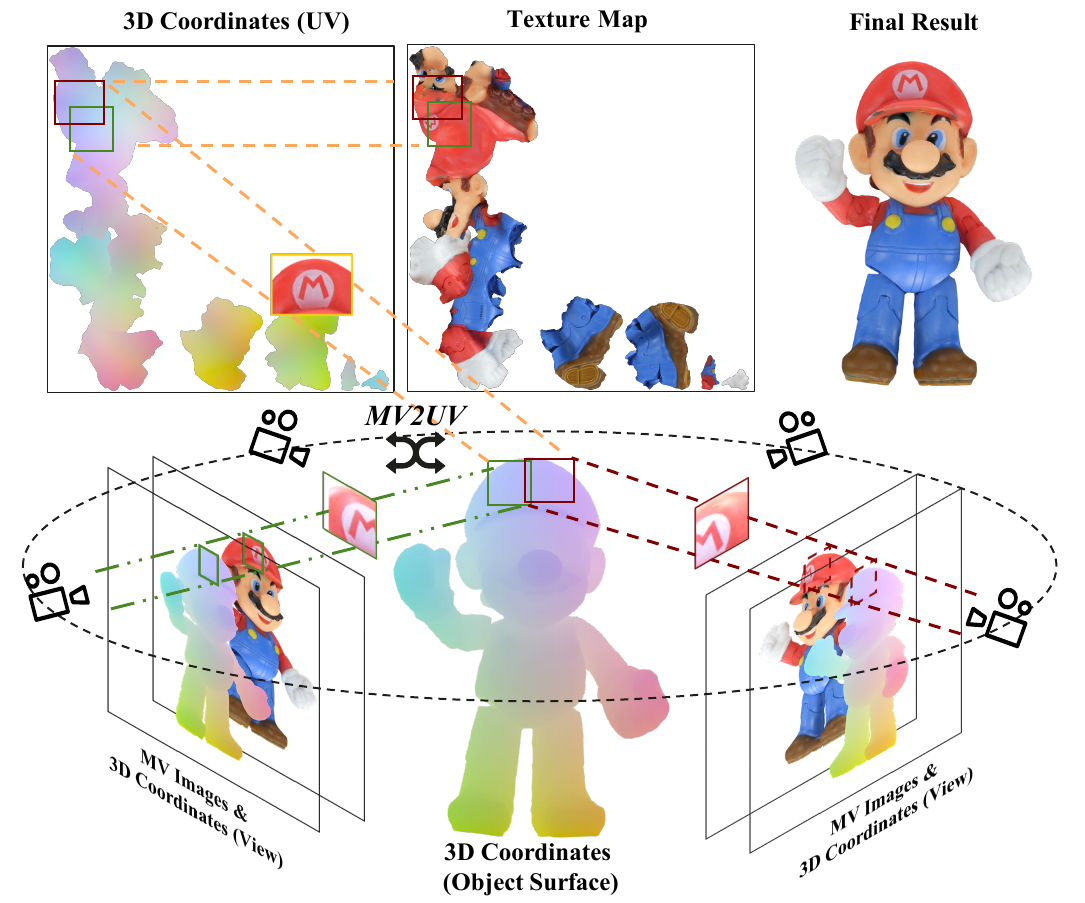}
    \caption{We apply the 3D coordinates as the positional encoding within MV-UV cross attention, enabling the UV map to attend to geometrically corresponding regions on multiview images for improved inpainting and inconsistency resolution.}
    \label{fig:motivation}
\end{figure}

\subsubsection{Encoding Multiview Conditions}
\label{subsec:mv}
This section aims to encode the generated multiview images and use encoded features as conditions to guide UV texture map generation. We begin with the feature extraction from multiview images using a reference network.

\textbf{Multiview Reference Network.}
To extract features from multiview images as prompts, we copy the denoising U-Net of SDXL as a reference network with frozen weights. We feed the VAE encoded latents of each input view into this reference network, setting the timestep $t=0$ to preserve original image information without introducing noise. The resulting feature maps of each block are injected into the UV diffusion model via reference attention layers.

\textbf{Reference Attention Layer.}
To integrate information from multiview images for UV texture generation, we employ a reference attention layer to correlate the generated UV map with the multiview image features by
\begin{align}
\begin{split}
\text{ViewRefAttn}&(h_{\text{in}}, f_{\text{view}}, p_{\text{uv}}, p_{\text{view}}) \\
&= \text{Softmax}\left( \frac{Q_{\text{ref}} \cdot K_{\text{ref}}^T}{\sqrt{d}} \right) V_{\text{ref}}, \\
Q_{\text{ref}} &= \text{Linear}(h_{\text{in}}) + p_{\text{uv}}, \\
K_{\text{ref}} &= \text{Linear}(f_{\text{view}}) + p_{\text{view}}, \\
V_{\text{ref}} &= \text{Linear}(f_{\text{view}}),
\end{split}
\end{align}
where $h_{in} \in \mathbb{R}^{(H \times W) \times C}$ denotes the feature map of the UV map with spatial dimensions $H \times W$, and $f_{view} \in \mathbb{R}^{(H' \times W' \times N) \times C}$ represents the extracted multiview features from $N$ views, each of spatial size $H' \times W'$. 

An essential modification here is that we adopt the geometric position encoding $p_{uv}$ and $p_{view}$ in the queries and keys to allow the attention layer to focus on relevant regions on the multiview images. We illustrate how we compute these two terms in the following.

\begin{figure*}[h]
    \centering
    \includegraphics[width=0.9\linewidth]{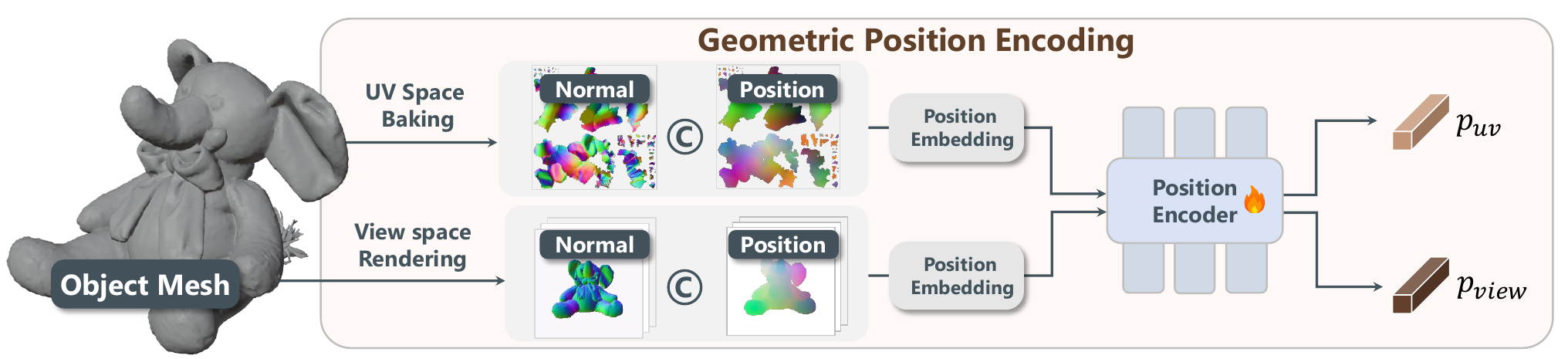}
    \caption{The geometric position encoding. We first render the object mesh to obtain normal and position maps, which are processed in two separate branches: one in UV space and the other in view space. In each branch, the corresponding geometric maps are transformed into positional embeddings and further encoded by a learnable position encoder to produce geometric features.}
    \label{fig:pe}
\end{figure*}

\subsubsection{Geometric Position Encoding}
\label{subsec:positionembedding}

For every pixel of the UV map and the multiview images, the corresponding 3D coordinates of these pixels indicate their positions on the 3D object. As shown in Fig.~\ref{fig:motivation} and Fig.~\ref{fig:pe}, to generate the color for a specific pixel on the UV map with a specific 3D coordinates, we need this pixel to refer to the corresponding pixels on the multiview images with similar 3D coordinates. Thus, we adopt their corresponding 3D coordinates as the positional encoding in the attention layers to improve such spatial awareness.

\textbf{Position Embedding.} 
We first apply Fourier function-based positional embedding~\cite{2020NeRF} on these 3D coordinates to better capture high-frequency signals. Note that the normal directions also indicate the local geometric information about the 3D shapes, so we concatenate them with the 3D coordinates before applying the Fourier positional embedding, which is then fed into a learnable position encoder.

\textbf{Learnable Position Encoder.} We introduce a learnable position encoder that transforms the initial position embeddings into a pyramid of features through a series of convolutional residual blocks. These features have the shapes aligned with the corresponding features in the corresponding attention layers, which are denoted by $p_{uv}$ and $p_{view}$. Both multiview images and UV maps share the same learnable position encoder.

\subsubsection{UV Self Attention}
\label{subsec:uv}

\begin{figure}
    \centering
    \includegraphics[width=0.9\linewidth]{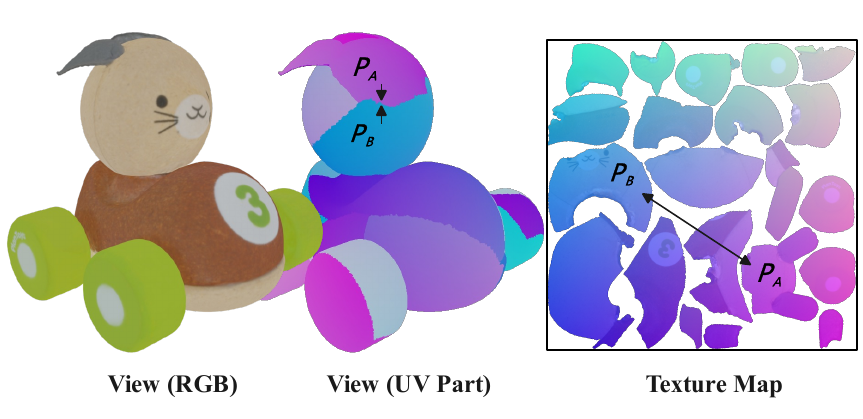}
    \caption{The UV Islands. UV self attention enables the model to establish correlations between UV regions (e.g., $P_A$ and $P_B$) that are adjacent in 3D space, even when they are disconnected in the UV layout.}
    \label{fig:uvisland}
\end{figure}

With the above geometric encoding, we further introduce a UV self attention layer to improve the spatial awareness on UV maps. As shown in Fig.~\ref{fig:uvisland}, standard image diffusion models struggle with the spatial discontinuity of UV islands in texture maps. Thus, we propose a new UV self attention layer to improve the spatial awareness so that the UV island can attend to other UV regions that are closer to it in 3D. This UV self attention layer incorporates the above geometric positional encoding to achieve this goal, which can be represented as follows:
\begin{align}
\begin{split}
\text{UVSelfAttn}&(h_{\text{in}}, p_{\text{uv}}) \\
&= \text{Softmax}\left( \frac{Q_{\text{uv}} \cdot K_{\text{uv}}^T}{\sqrt{d}} \right) V_{\text{uv}}, \\
Q_{\text{uv}} &= \text{Linear}(h_{\text{in}}) + p_{\text{uv}}, \\
K_{\text{uv}} &= \text{Linear}(h_{\text{in}}) + p_{\text{uv}}, \\
V_{\text{uv}} &= \text{Linear}(h_{\text{in}}),
\end{split}
\end{align}
By adding position embedding $p_{uv}$ in both the queries and keys, the model can easily learn stronger attention weights between pixels closer in the 3D space. This bias encourages the model to attend to features based on their 3D geometric relationships, thereby effectively bridging the gaps between isolated UV islands and fostering the synthesis of a globally consistent texture.

\subsubsection{Architecture Summary}
\label{subsec:summary}
With the above three new designs, we summarize our modification on the architecture as follows.
As illustrated in the right panel of Fig.~\ref{fig:pipeline}, we employ a parallel attention mechanism to integrate reference attention and UV self attention. Specifically, we copy the weights of the original self attention layers from the frozen SDXL backbone to construct these two new attention layers. This block can be formulated as
\begin{align}
\begin{split}
h_{out} &= h_{\text{in}} + \text{SelfAttn}(h_{\text{in}}) \\
       &\quad + \text{ViewRefAttn}(h_{\text{in}}, f_{\text{view}}, p_{\text{uv}}, p_{\text{view}}) \\
       &\quad + \text{UVSelfAttn}(h_{\text{in}}, p_{\text{uv}}).
\end{split}
\end{align}
This design ensures that the network retains the pre-trained priors encoded in the original self attention layers. Meanwhile, the parallel reference attention layer and UV self attention layer enable the model to learn task-specific capabilities: leveraging multiview images as prompts to guide UV texture generation, and modeling intrinsic relationships within the UV space. Together, these parallel modules extend the backbone’s functionality to directly synthesize UV maps while preserving the strengths of the pre-trained model.

\begin{figure*}
    \centering 
    \includegraphics[width=0.98\textwidth]{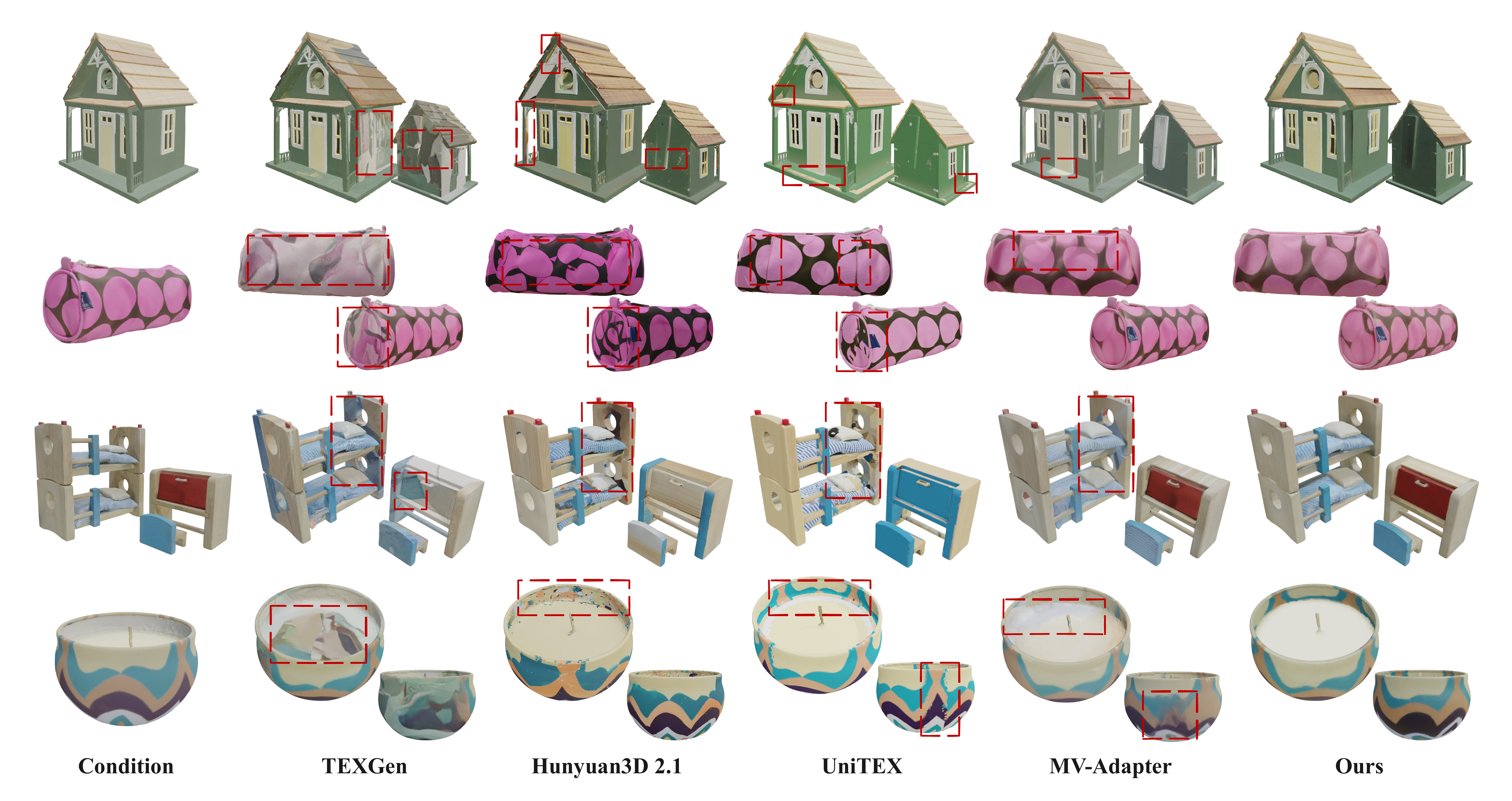}  
    \caption{Comparisons with texture generation methods from a single image input. Our method enables automatic completion of textures in occluded UV regions while avoiding inconsistencies induced by direct projection.} 
    \label{fig:comparewithsota}
\end{figure*}
\section{Experiment}
\textbf{Datasets.} We trained our model on the dataset provided by Material Anything \cite{huang2025material}, a subset of Objaverse ~\cite{deitke2023objaverse} comprising approximately 80,000 samples. For objects with multiple parts, all components were merged into a single mesh, and a new UV mapping is generated via the X-atlas Project Tool\cite{xatlas}. For each model, we rendered albedo, position, and global normal from 6 fixed viewpoints and baked them into UV maps. To enhance the model’s robustness under varying lighting conditions, diverse illumination settings—including point lights, ambient lights, and area lights—were randomly applied during rendering. Meanwhile, to address multiview inconsistencies, the MV-Adapter model was employed to perform img-to-img redrawing on rendered views with strengths of [0.1, 0.25, 0.5]. Both types of processed data were randomly sampled during training with a probability of 20\% as a data augmentation strategy. To balance performance and efficiency, we set the multiview resolution to $768 \times 768$ and the texture map resolution to $1024 \times 1024$. Additional implementation details are provided in the supplementary material.

\begin{figure}  
    \centering       
    \includegraphics[width=0.98\linewidth]{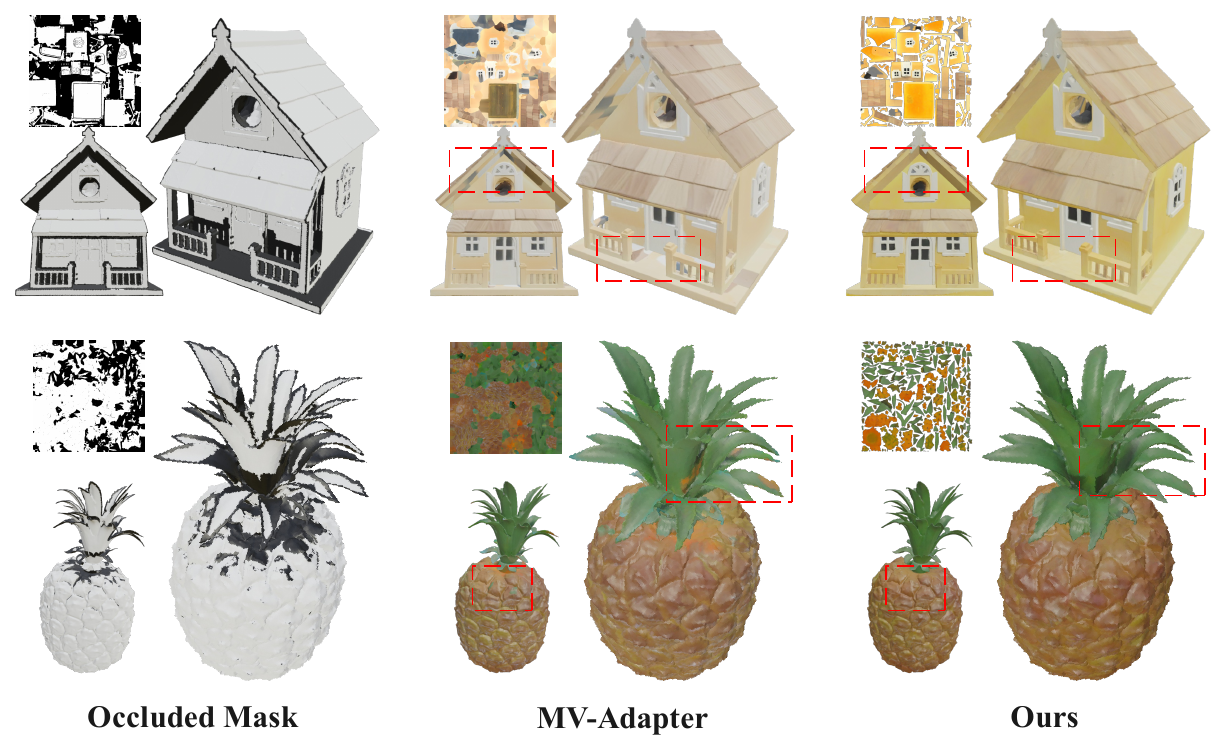}
    \caption{Qualitative comparison on occluded regions. The occlusion mask (left), MV-Adapter results (middle), and our results (right) demonstrate our method's capability in resolving view inconsistencies and producing more semantically coherent results.}  
    \label{fig:occludedfid}  
\end{figure}

\subsection{Image-Conditioned Texture Generation}
For a systematic evaluation, we conducted a comparative study between our method and state-of-the-art texture generation methods, including TEXGen, UniTEX, Hunyuan3D 2.1 and MV-Adapter. We randomly sampled 200 instances from both the Google Scanned Objects (GSO) dataset ~\cite{downs2022google} and the Digital Twin Catalog (DTC) dataset ~\cite{dong2025digital}, utilizing a rendered image from a fixed viewpoint as the reference image for each method. To evaluate the end-to-end generation capability of our method, we first employ MV-Adapter to generate MVs, which are then used as inputs of our network to synthesize the final textures.
\begin{table}
    \centering
    \caption{Performance comparison of material generation methods. Our approach achieves the best FID and KID scores on both datasets, demonstrating superior performance.}
    \label{tab:comparewithsota}
    \begin{threeparttable}
        \begin{tabular*}{\linewidth}{@{\extracolsep{\fill}}lcccc@{}}
            \toprule
            \multirow{2}{*}{Method} & \multicolumn{2}{c}{GSO Dataset} & \multicolumn{2}{c}{DTC Dataset} \\
            \cmidrule(lr){2-3} \cmidrule(lr){4-5}
            & FID ($\downarrow$) & KID ($\downarrow$) & FID ($\downarrow$) & KID ($\downarrow$) \\
            \midrule
            TEXGen & 75.2 & 283.3 & 41.1 & 118.6 \\
            Hunyuan3D 2.1 & 30.8 & 72.5 & 41.4 & 49.9 \\
            UniTEX & 39.2 & 101.5 & 39.7 & 43.3 \\
            MV-Adapter & 24.7 & 47.5 & 28.4 & 41.8 \\
            Ours & 24.4 & 43.6 & 26.4 & 28.7 \\
            \bottomrule
        \end{tabular*}
    \end{threeparttable}
\end{table}

Quantitative results in Tab.~\ref{tab:comparewithsota} show that our method achieves the best Fréchet Inception Distance (FID) and Kernel Inception Distance (KID) scores ($\times 10^{-4}$) on both GSO and DTC datasets, substantially outperforming existing approaches. Fig.~\ref{fig:comparewithsota} further demonstrates that our method produces more complete textures with fewer artifacts from the same input. Compared with the UV-based TEXGen, our results exhibit richer details and more plausible structure. Against multiview-based methods (e.g., Hunyuan3D 2.1, UniTEX, MV-Adapter), our approach significantly reduces view-inconsistency artifacts and improves semantic coherence, particularly in occluded regions.

\subsection{Ablation Studies}
In this section, we assess the semantic completion capability of our method on occluded regions, evaluate its ability to resolve view inconsistencies, and conduct comprehensive ablation studies to analyze the effect of each key component.

\textbf{Occluded Region Inpainting.}  
We evaluate occluded region inpainting on 10 self-occluded samples using zoomed-in renders. Our method significantly outperforms MV-Adapter (Tab.~\ref{tab:occluded_evaluation}), reducing FID by 56.1 and KID by 110.3 through effective fusion of multiview semantics and UV-based generation. Fig.~\ref{fig:occludedfid} qualitatively shows our approach resolves view inconsistencies and produces semantically coherent completions in occluded UV regions.

\begin{table}[t]
    \centering
    \caption{Quantitative evaluation on occluded regions. FID and KID are computed between rendered images and the ground truth forced on occluded part. Our method achieves superior performance over the MV-Adapter baseline. }
    \label{tab:occluded_evaluation}
    \begin{threeparttable}
        \begin{tabular}{lcc}
            \toprule
            Method & FID ($\downarrow$) & KID ($\downarrow$) \\
            \midrule
            MV-Adapter & 123.6 & 157.9 \\
            Ours & 67.5 & 47.6 \\
            \bottomrule
        \end{tabular}
    \end{threeparttable}
\end{table}

\begin{figure}  
    \centering       
    \includegraphics[width=1.0\linewidth]{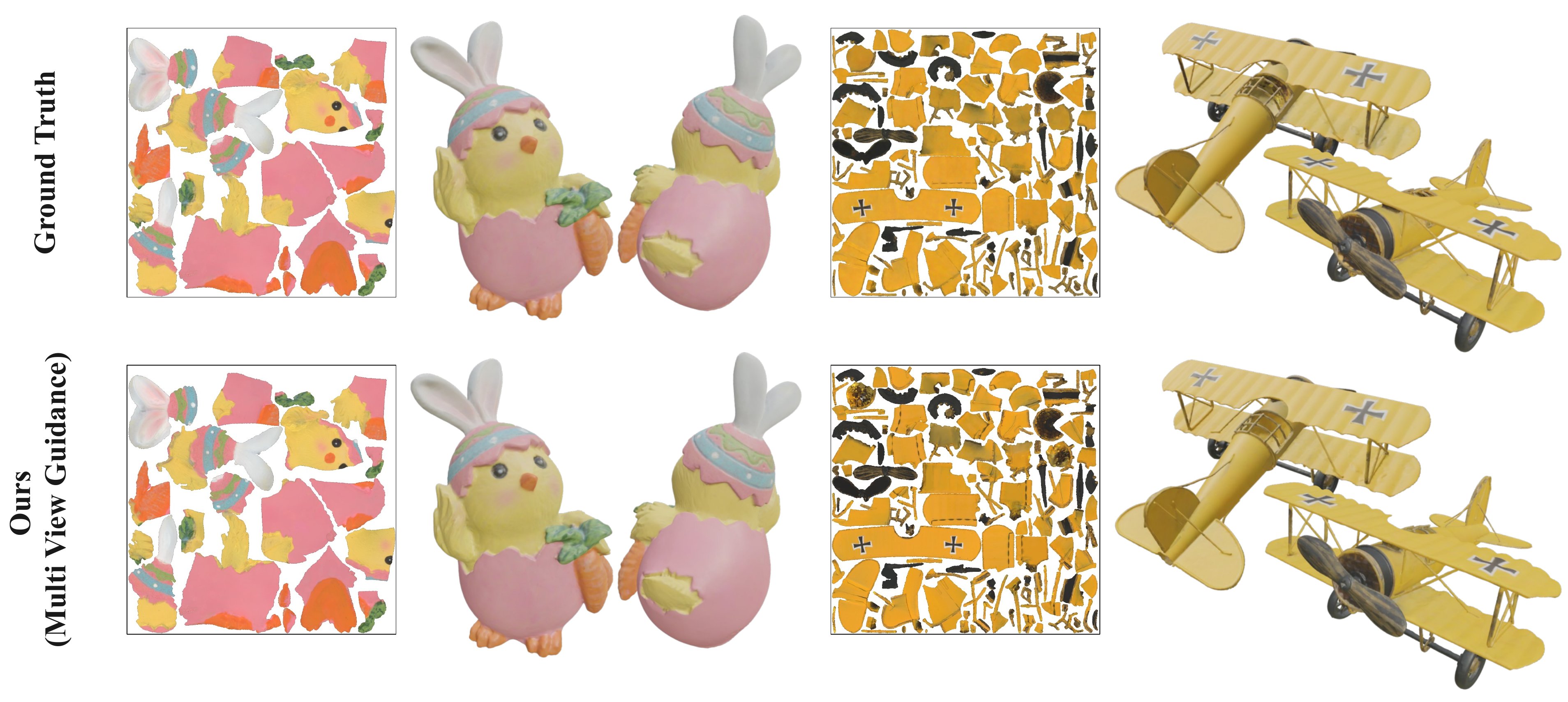}
    \caption{The generated results with rendered views.}  
    \label{fig:rebuild}  
\end{figure}

\textbf{View-inconsistency Resolution.} Our framework is designed to autonomously reconcile potential view conflicts in UV-based generation, thereby preventing the artifacts that arise from direct projection. We first validate our method's ability to link multiview semantics to UV space using 6 fixed viewpoints rendered from the models in the GSO and DTC datasets. As shown in the GSO and DTC rows of Tab.~\ref{tab:rebuild} and Fig.~\ref{fig:rebuild}, our method achieves faithful texture reconstruction (PSNR/SSIM against ground truth) and high-quality novel view synthesis (FID/KID on 26 surrounding views).

To specifically assess view inconsistency resolution, we synthetically create input conflicts by replacing the original back view with an image regenerated by MV-Adapter with strength of 0.5. As shown in the $GSO_c$ and $DTC_c$ rows of Tab.~\ref{tab:rebuild}, our method maintains robust reconstruction quality despite these inconsistencies, with only marginal metric degradation: PSNR decreases by 0.2 on both datasets, while SSIM drops by merely 0.003 on GSO and 0.001 on DTC. This minimal performance loss demonstrates our method's effective handling of view conflicts.

\begin{table}
    \centering
    \small
    \caption{Ablation study on View-inconsistency Resolution.}
    \begin{tabular}{lcccc}
        \toprule
        \multirow{2}{*}{Dataset} & \multicolumn{4}{c}{Metrics} \\
        \cmidrule(lr){2-5}
         & PSNR $\uparrow$ & SSIM $\uparrow$ & FID $\downarrow$ & KID $\downarrow$ \\
        \midrule
        GSO    & 25.7 & 0.855 & 3.32 & 8.49 \\
        $GSO_{c}$    & 25.5 & 0.852 & 4.03 & 10.6 \\
        DTC    & 28.8 & 0.886 & 5.44 & 14.9 \\
        $DTC_{c}$     & 28.6 & 0.885 & 5.90 & 16.4 \\
        \bottomrule
    \end{tabular}
    \label{tab:rebuild}
\end{table}

\begin{figure}       
    \centering       
    \includegraphics[width=0.96\linewidth]{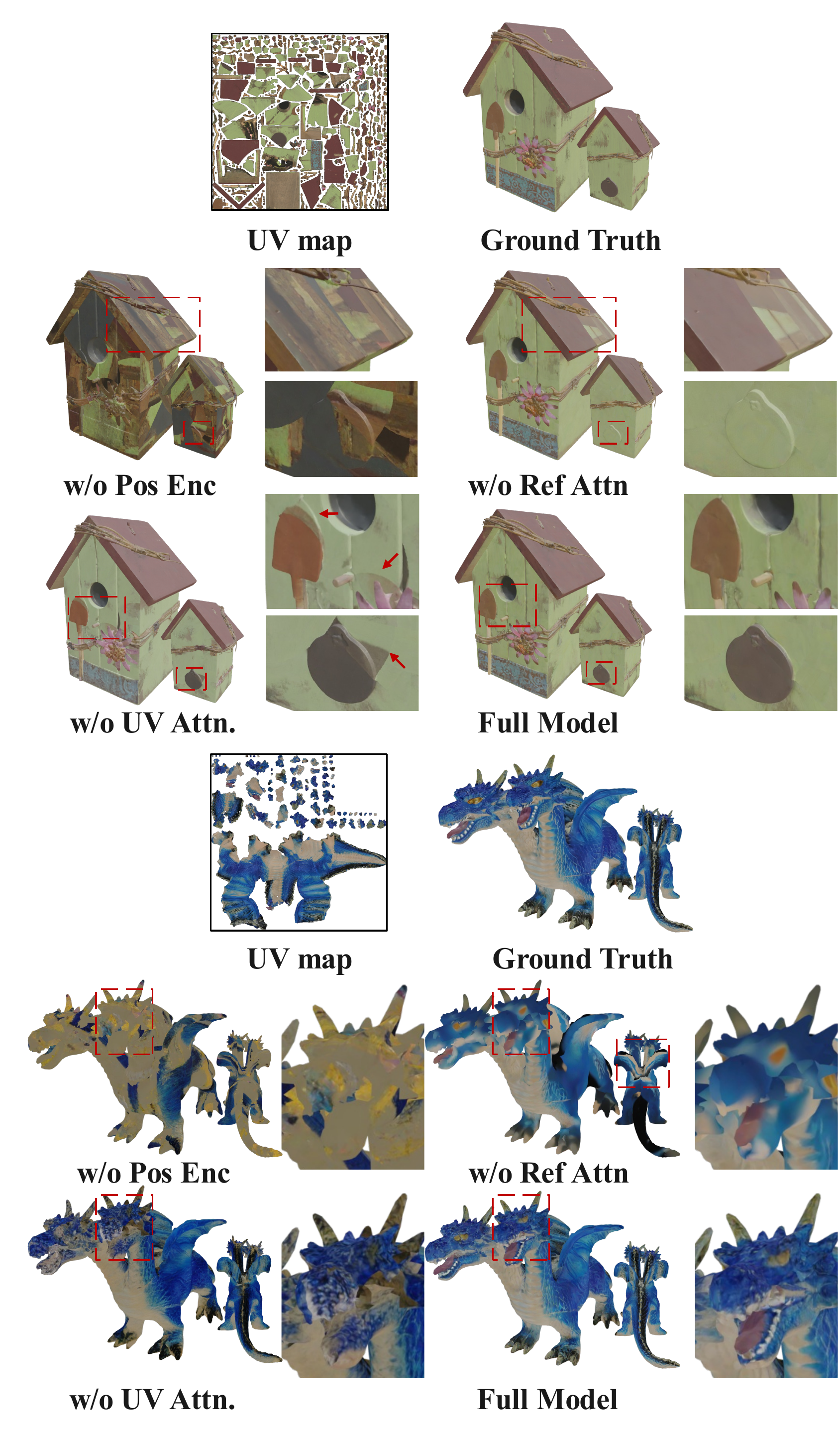}
    \caption{Ablation study results}  
    \label{fig:ablation}  
\end{figure}

\textbf{Reference Attention Layer.}
The core design of our framework leverages multiview images as semantic prompts for UV-space texture generation. To validate the effectiveness of the proposed Reference Attention mechanism, we ablate this component and adopt a minimal inpainting configuration following Stable Diffusion~\cite{rombach2022high}, where the front view is back-projected and concatenated with latent noise. As shown in Tab.~\ref{tab:ablation_study}, the ablated model exhibits significant performance degradation across all metrics. Fig.~\ref{fig:ablation} further illustrates that while basic texture completion is achieved, the model fails to generate semantically coherent textures without Reference Attention.

\textbf{Geometric Position Encoding.} To investigate the importance of geometric position encoding, we conduct an ablation study. We remove only the geometric position encoding while retaining all other attention mechanisms. As shown in Tab.~\ref{tab:ablation_study} and Fig.~\ref{fig:ablation}, although the network can still capture some high-level color features through the Reference Attention mechanism, the lack of uniquely cross-modal 3D coordinates guidance prevents it from focusing on semantically coherent regions, resulting in significant artifacts.

\textbf{UV Self Attention.} To evaluate the effectiveness of the UV self attention module, we perform an ablation study by removing this component. Quantitative results in Tab.~\ref{tab:ablation_study} confirm that its absence degrades spatial coherence in generated textures. As visually demonstrated in Fig.~\ref{fig:ablation}, the module enables geometric awareness across UV islands—without it, the model fails to maintain texture continuity between adjacent 3D regions, leading to semantic inconsistencies. This limitation is especially pronounced in structurally complex areas such as the dragon's head.

\begin{table}
    \centering
    \caption{Ablation study on GSO and DTC datasets. Configurations: without geometric position encoding (w/o Pos Enc), reference attention (w/o Ref Attn), UV self attention (w/o UV Attn).}
    \label{tab:ablation_study}
    \begin{tabular}{lcccc}
        \toprule
        Configuration & \multicolumn{2}{c}{GSO Dataset} & \multicolumn{2}{c}{DTC Dataset} \\
        \cmidrule(lr){2-3} \cmidrule(lr){4-5}
        & PSNR $\uparrow$ & SSIM $\uparrow$ & PSNR $\uparrow$ & SSIM $\uparrow$ \\
        \midrule
        w/o Pos Enc & 13.6 & 0.688 & 13.9 & 0.689 \\
        w/o Ref Attn & 18.9 & 0.805 & 22.6 & 0.861 \\
        w/o UV Attn & 21.5 & 0.814 & 23.9 & 0.863 \\
        Full Model & 25.7 & 0.855 & 28.8 & 0.886 \\
        \bottomrule
    \end{tabular}
\end{table}

\subsection{Limitation}
Our method effectively captures fine-grained texture details, yet struggles with textual elements due to resolution constraints and VAE capacity, inevitably introducing distortion. We observe that even when reconstructing the texture map directly through the VAE provided by Stable Diffusion XL, the text regions still exhibit noticeable blurring. This observation reveals a limitation of our current method. We identify resolving this challenge as a valuable direction for future research.

\section{Conclusion}
In this paper, we propose a novel framework that bridges 2D generative priors from multiview generations with UV-space generation. Our approach treats multiview images as semantic prompts to guide UV texture generation through a diffusion model rather than direct back-projection. Quantitative and qualitative results demonstrate its superiority in both conflict resolution and semantic inpainting in occluded region.
  
{
    \small
    \bibliographystyle{ieeenat_fullname}
    \bibliography{main}
}


\let\twocolumn\oldtwocolumn
\clearpage
\setcounter{page}{1}
\maketitlesupplementary

\section{Training Details}
\label{sec:training}
We finetune our backbone network based on Stable Diffusion XL (SDXL) with frozen self attention and cross attention modules. To enhance the model’s ability to complete UV textures, we randomly drop each view with a probability of 0.1. We do not employ classifier-free guidance (CFG). Additionally, we shuffle the order of input multiview images as a form of data augmentation. The model is trained for 10 epochs on 80K samples with a learning rate of 5e-5 and a batch size of 32.

\section{Detail of Geometric Position Encoding}
\label{sec:positionencoder}

\begin{figure*}
    \centering
    \includegraphics[width=0.8\linewidth]{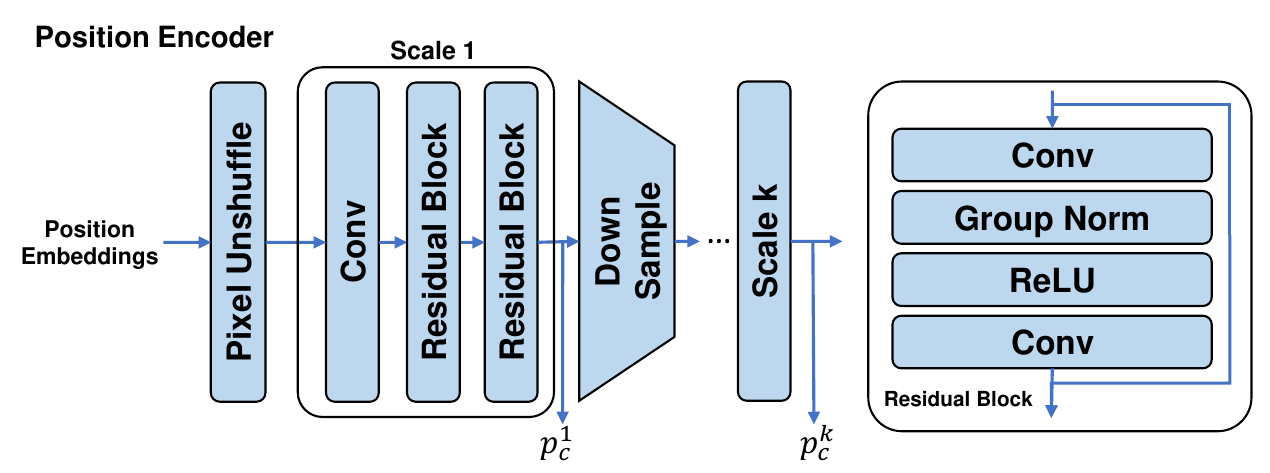}
    \caption{Detail of Learnable Position Encoder.}
    \label{fig:positionencoder}
\end{figure*}

\begin{figure*}
    \centering
    \includegraphics[width=1.0\linewidth]{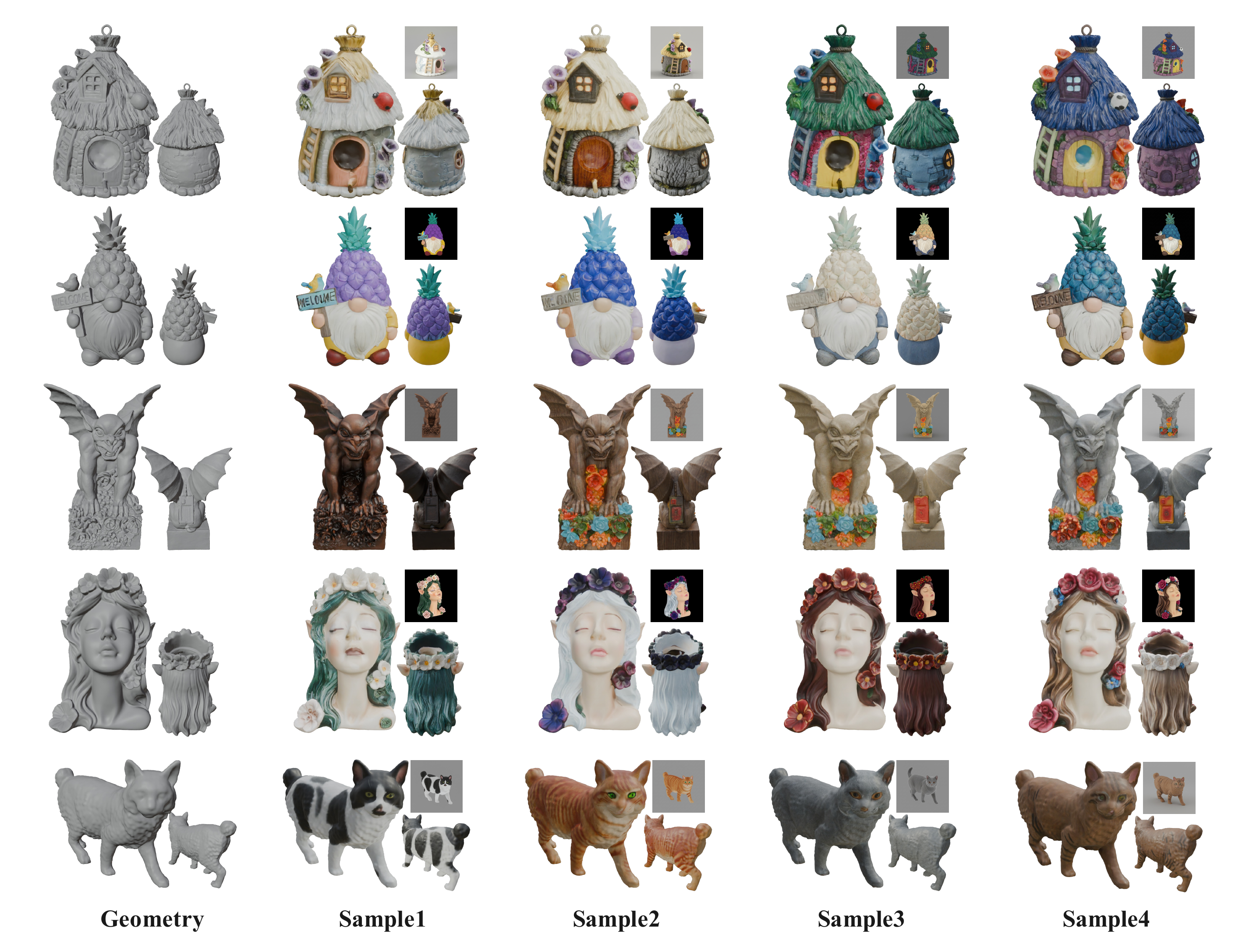}
    \caption{MV2UV can generate diverse, high-quality textures from different prompts, enabling flexible and controllable texture synthesis.}
    \label{fig:multisample}
\end{figure*}

\textbf{Position Embedding.}
We normalize both the position $(x, y, z)$ and normal $(n_x, n_y, n_z)$ maps to the range $[-1, 1]$ and concatenate them along the channel dimension to form the input 3D coordinates representation. To better capture high-frequency geometric signals, we apply a Fourier-based positional embedding to these coordinates, defined as:
\[
\begin{split}
\gamma(p) = [ & \sin(2_0 \pi p),\ \cos(2_0 \pi p),\ \ldots, \\
              & \sin(2_{L-1} \pi p),\ \cos(2_{L-1} \pi p) ]
\end{split}
\]
where each dimension $p$ of the input coordinates is projected into a series of sinusoids. The hyperparameter $L$ controls the highest frequency band and thus the fidelity of representable geometric details. We set $L = 10$ in all experiments.\\
\textbf{Learnable Position Encoder.} We further introduce a learnable position encoder that transforms the initial position embeddings into a multi-scale feature pyramid via a series of convolutional residual blocks. As shown in \cref{fig:positionencoder}, the input embedding $p_{\text{in}} \in \mathbb{R}^{H \times W \times C}$ is first downsampled via a Pixel Unshuffle layer, rearranging it into a lower-resolution feature map of size $\mathbb{R}^{\frac{H}{8} \times \frac{W}{8} \times (C \times 8 \times 8)}$. At each scale, a convolutional layer followed by two residual blocks extracts positional features $p_c^k$. The final output is a pyramid of features $[p_c^1, \dots, p_c^k]$ whose spatial dimensions align with the corresponding feature maps in the diffusion model’s attention layers. This design enables fine-grained control over the UV texture generation process, facilitating high-precision detail synthesis across multiple spatial scales.

\section{Diverse Texturing}
\label{sec:diversetextureing}
Given an untextured 3D mesh, we first render a single view of the model and apply Nano Banana to perform example-based image generation, producing a reference image. This reference image is then fed into MV2UV to generate the corresponding texture map. As shown in Fig.~\ref{fig:multisample}, our approach can produce diverse, high-quality textures conditioned on different image prompts, enabling flexible and controllable texture synthesis for 3D shapes.

\end{document}